\documentclass[sigconf]{acmart}
\usepackage{booktabs} 
\usepackage{natbib}
\usepackage{graphicx}
\usepackage{amsmath,bm,amsfonts,hyperref}
\usepackage{caption}
\usepackage{subfigure}
\usepackage[para,online,flushleft]{threeparttable}
\usepackage{multirow}
\usepackage{makecell}
\setcopyright{acmcopyright}

\begin{document}
\title{Fault Sneaking Attack: a Stealthy Framework for Misleading Deep Neural Networks}




\author{Pu Zhao, Siyue Wang, Cheng Gongye, Yanzhi Wang, Yunsi Fei,  Xue Lin}
\affiliation{%
  \institution{Northeastern University, Boston, Massachusetts}
  \city{ \{zhao.pu, wang.siy, gongye.c\}@husky.neu.edu,  yanz.wang@northeastern.edu,\\ yfei@ece.neu.edu, xue.lin@northeastern.edu }
    \country{USA}
}

\begin{abstract}
Despite the great achievements of deep neural networks (DNNs), the vulnerability of state-of-the-art DNNs raises security concerns of DNNs in  many application domains requiring high reliability. 
We propose the fault sneaking attack on DNNs, where the adversary aims to misclassify certain input images into any target labels by modifying the DNN parameters. 
We apply ADMM (alternating direction method of multipliers) for solving the optimization problem of the fault sneaking attack with two constraints: 1) the classification of the other images should be unchanged 
and 2) the parameter modifications should be minimized. 
Specifically, the first constraint requires us not only to inject designated faults (misclassifications), but also to hide the faults for stealthy or sneaking considerations by maintaining model accuracy.
The second constraint requires us to minimize the parameter modifications (using $\ell_0$ norm to measure the number of modifications and $\ell_2$ norm to measure the magnitude of modifications).
Comprehensive experimental evaluation demonstrates that the proposed  framework can inject multiple sneaking faults without losing the overall test accuracy performance.
\end{abstract}

%

\ccsdesc[500]{Security and privacy~Domain-specific security and privacy architectures}
\ccsdesc[300]{Security and privacy~Network security}
\ccsdesc[300]{Networks~Network performance analysis}
\ccsdesc[300]{Theory of computation~Theory and algorithms for application domains}

\keywords{Deep neural networks, Fault injection, ADMM}

\copyrightyear{2019} 
\acmYear{2019} 
\setcopyright{acmcopyright}
\acmConference[DAC '19]{The 56th Annual Design Automation Conference 2019}{June 2--6, 2019}{Las Vegas, NV, USA}
\acmPrice{15.00}
\acmDOI{10.1145/3316781.3317825}
\acmISBN{978-1-4503-6725-7/19/06}

\maketitle

\section{Introduction}

Modern technologies based on pattern recognition, machine learning, and specifically deep learning, have achieved significant breakthroughs \cite{lecun2015deep} in a variety of application domains. 
Deep neural network (DNN) has become a fundamental element and a core enabler in the ubiquitous artificial intelligence techniques. 
However, despite the impressive performance, many recent studies demonstrate that state-of-the-art 
DNNs are vulnerable to adversarial attacks \cite{goodfellow2015explaining,szegedy2013intriguing}.
This raises concerns of the DNN robustness in many applications with high reliability and dependability requirements such as face recognition, autonomous driving, and malware detection \cite{mahmood2017adversarial,evtimov2017robust}. 

After the exploration of adversarial attacks in image classification and objection detection from 2014, the vulnerability and robustness of DNNs have attracted ever-increasing attentions and efforts in the research field known as \textit{adversarial machine learning}.  
Since then, a large amount of efforts have been devoted to: 1) design of adversarial attacks against machine learning tasks \cite{carlini2017towards,chen2017ead,zhao2018admm,pu2019admm};
2) security evaluation methodologies to systematically estimate the DNN robustness \cite{biggio2014security,zhang2018efficient,Weng_Zhao_Liu_Chen_Lin_Daniel_2020,ijcai2021p80,osti_10300680,8646651};
and 3) defense mechanisms under the attacks \cite{rota2017randomized,demontis2018yes,madry2017towards,zhao2020bridging,8646578}. This paper falls into the first category.

The adversarial attacks can be classified into: 1) evasion attacks \cite{carlini2017towards,chen2017ead,zhao2018admm,Zhao_Chen_Wang_Lin_2020,9009069,xu2018structured} that perturb input images at test time to fool DNN classifications; 2) poisoning attacks \cite{xiao2015feature,biggio2012poisoning} that manipulate training data sets to obtain illy-trained DNN models; 
and 3) fault injection attacks \cite{liu2017fault,breier2018practical} that change classifications of certain input images to the target labels by modifying DNN parameters.
The general purpose of an adversarial attack no matter its category is to have misclassifications of certain images, while maintaining high model accuracy for the other images. 
This work proposes the fault sneaking attack, a new method of the fault injection attack.

Fault injection attack perturbs the DNN parameter space.
As DNNs are usually implemented and deployed on various hardware platforms including CPUs/GPUs and dedicated accelerators, it is possible to perturb the DNN parameters stored in memory enabled by the development of memory fault injection techniques such as laser beam \cite{selmke2015precise} and row hammer \cite{kim2014flipping}. 
To be practical, we propose the fault sneaking attack to perturb the DNN parameters with considerations of attack implementation in the hardware.

It is a more challenging task to perturb the parameters (as fault injection attack) than to perturb the input images (as evasion attack) due to the following two reasons: 1) global effect: perturbing one input would not influence the classifications of other unperturbed inputs while perturbing the parameters has a global effect for all inputs; 2) numerous parameters: the DNNs usually have a much greater number of parameters than the pixel number of an input image. 
The fault injection attack should be stealthy in that misclassifications are only for certain images while maintaining high model accuracy for the other images, and therefore cannot be easily detected. And it should also be efficient in that the parameter modifications should be as small as possible, and therefore can be implemented easily in the hardware.
This work tackles these challenges by proposing the fault sneaking attack based on ADMM (alternating direction method of multipliers).


The \textbf{theoretical} contributions of this work are:

+ \emph{\textbf{Stealthy injection of multiple faults:}} The proposed fault sneaking attack based on ADMM enables to achieve multiple designated faults (misclassifications) with the flexibility to specify any target labels and the stealthiness to hide the faults. The fault injection attack \cite{liu2017fault} can only inject one fault.   



+ \emph{\textbf{A systematic application of ADMM with analytical solutions:}} 
Comparing with the heuristic \cite{liu2017fault}, the proposed fault sneaking attack is an optimization based framework leveraging ADMM with analytical solutions. Comparing with evasion attacks \cite{zhao2018admm,carlini2017adversarial}, the proposed fault sneaking attack deals with a more challenging problem with higher dimensionality, but surprisingly finds much less expensive analytical solutions.


+ \emph{\textbf{A general ADMM framework for both $\ell_0$ and $\ell_2$ norm minimizations:}}  
The proposed ADMM based framework for solving the optimization problem of fault sneaking attack can adopt both $\ell_0$ norm (the number of parameter modifications) and $\ell_2$ norm (the magnitude of modifications) to measure the difference between original and modified DNN models with only minor changes. However, \cite{liu2017fault} cannot deal with the non-differential $\ell_0$ norm.  



The \textbf{experimental} contributions of this work are:

+ \emph{\textbf{Less model accuracy loss:}}
Under the same experimental settings and misclassification requirements, the proposed fault sneaking attack degrades the DNN model accuracy by only 0.8 percent for MNIST and 1.0 percent for CIFAR, while \cite{liu2017fault} degrades the DNN model accuracy by 3.86 percent and 2.35 percent, respectively.


+ \emph{\textbf{Comprehensive analysis of DNN fault tolerance:}}
We extensively test the capability of DNNs on tolerance of fault injection attacks. We find that there is an upper limit on the number $S$ of images with successful misclassifications 
depending on the DNN model itself. For the DNN models used in this work, the number $S$ is around 10 demonstrating the tolerance for sneaking faults as 10.

\section{Related Work}


The adversarial attacks are reviewed from the aspects of perturbing the inputs and perturbing the DNN parameters.

\subsection{Perturbing the Input Space}
Evasion attacks generate adversarial examples to fool DNNs by perturbing the legitimate inputs.
Basically, an adversarial example is produced by adding human-imperceptible distortions onto a legitimate image, such that the adversarial example will be classified by the DNN as a target (wrong) label.
The  norm-ball  constrained  evasion  attacks have been well studied, including  the
FGM \citep{goodfellow2014explaining} and IFGSM  \citep{KurakinGB2016adversarial} attacks with $\ell_\infty$ norm restriction, the L-BFGS  \citep{szegedy2013intriguing} and C\&W \citep{carlini2017towards} attacks minimizing the $\ell_2$  distortion, and the JSMA \citep{papernot2016limitations} and ADMM \citep{pu2018reinforced} attacks trying to perturb the minimum number of pixels, namely,  minimizing the $\ell_0$ distortion. 

Many defense works have been proposed, including defensive distillation \citep{papernot2016distillation} 
,  defensive dropout \cite{wang2018defending,wang2018defensive}
, and robust adversarial training.  \citep{madry2017towards} 
The robust adversarial training method ensures strong defense performance with high computation requirement. 

\subsection{Perturbing the Parameter Space}
Poisoning attacks, which train DNNs by adding  poisoned images 
into the training data sets, and fault injection attacks, which modify the DNN parameters directly, are attacks that perturb the DNN parameters.
Poisoning attack \cite{xiao2015feature} is  computation-intensive as it requires iterative retraining and is not the focus of our paper.
Fault injection attack \cite{liu2017fault} was first proposed by Liu et al, which uses a heuristic approach to profile the sink class 
for single bias attack scheme, and  compresses the modification by iteratively enforcing the smallest element as zero and feasibility check for gradient descent attack scheme.
Different from \cite{liu2017fault}, the fault sneaking attack uses a systematic optimization-based approach, achieving flexible designations of target labels and portion of DNN parameters to modify, and enabling both the $\ell_2$ and $\ell_0$ (non-differential) norms in the objective function.


\subsection{Practical Fault Injection Techniques}
The common techniques  flipping the logic values in memory include laser beam and row hammer. Laser beam \cite{barenghi2012fault} can precisely change any single bit in SRAM by carefully tuning the laser beam such as diameter and energy level \cite{selmke2015precise}. Row hammer \cite{kim2014flipping} can inject faults into DRAM by rapidly and repeatedly accessing a given physical memory location to flip corresponding bits \cite{xiao2016one}. Some works demonstrate the feasibility of using row hammer on mobile platforms \cite{van2016drammer} and launching the row hammer to trigger the processor lockdown \cite{jang2017sgx}. However, fine-tuning the laser beam or locating the bits in memory can be time consuming \cite{van2016drammer}. Therefore, it is essential to minimize the number of modified parameters by our fault sneaking attack.
Recently, \cite{breier2018practical} implements the DNN fault injection attack \cite{liu2017fault} physically on embedded systems using laser beam.
In particular, \cite{breier2018practical} injects faults into the widely used activation functions in DNNs and demonstrates the possibility to achieve misclassifications by injecting faults into the DNN hidden layer.

\section{Problem Formulation}



\textcolor{black}{\textbf{Threat Model:} We consider an adversary tampering with the DNN classification results of certain input images into designated target labels by modifying the DNN model parameters.
In this paper, we assume white-box attack, i.e., the adversary has the complete knowledge of the DNN model (including both structures and parameters) and low-level implementation details (how and where DNN parameters are located in the memory), as the highest and most stringent security standard to assess the robustness of DNN systems under fault sneaking attack.
Given existing fault injection techniques can precisely flip any bit of the data in memory, we assume the adversary can modify any parameter in DNN to any value that is in the valid range of the used arithmetic format.
Note that, we do not assume the adversary knows the training and testing data sets, which are usually not available to the system users. 
}

\textcolor{black}{The adversary has two constraints when launching the fault sneaking attack: (i) Stealthy, in that the classification results of the other images should be kept as unchanged as possible; (ii) Efficient, in that the modifications of DNN parameters in terms of number of modified parameters or magnitude of parameter modifications should be as small as possible.
The first constraint is important because even if the attack is specified for certain input images, it is highly possible to change the classification results of the other images when modifying the DNN parameters, thereby resulting in obviously low DNN model accuracy and easy detection of the attack.
The second constraint minimizing the parameter modifications can reduce the influence and difficulty of implementing the attack. 
}

\textbf{Attack Model:} Given $R$ images $\mathcal X = \left\{ {{\bm x_i}|i = 1, \cdots ,R} \right\}$ with their correct labels $\mathcal L = \{l_i | i = 1, \cdots ,R\}$, we would like to change the classification results of the first $S$ ($S\leq R$) images to their target labels $\mathcal T = \{t_i | i = 1, \cdots ,S\}$, while the classifications of the rest $R-S$ images are unchanged, by modifying parameters in the DNN model. Note that the unchanged labels of the other $R-S$ images are to make the attack stealthy and hard to detect.

The original DNN model parameters  are denoted as $\bm \theta$, and $\bm \delta$ represents the parameter modifications. 
So the parameters after the modification   are $\bm \theta + \bm \delta$.
Note that $\bm \theta$ has the flexibility of specifying either all the DNN parameters  or only a portion of the parameters, e.g., weight parameters of the specific layer(s).
The fault sneaking attack can be formulated as an optimization problem:
\begin{equation} \label{eq: original_problem}
\mathop {\min }\limits_{\bm \delta}  \quad D(\bm \delta ) + G(\bm \theta  + \bm \delta ,\mathcal X, \mathcal T, \mathcal L),
\end{equation}
where $D(\bm \delta )$ measures the DNN parameter modifications; and $G(\bm \theta + \bm \delta, \mathcal X, \mathcal T, \mathcal L)$ represents the misclassification requirements, i.e., with the modified DNN model parameters $\bm \theta + \bm \delta$, the first $S$ images in set $\mathcal X$ will be classified as target labels $\mathcal T$, \textcolor{black}{ while the classifications of the rest $R-S$ images are kept unchanged.} The details of the $D$ and $G$ functions are to be explained in the following sections. 


\subsection{Measurements of Parameter Modifications}
$D(\bm \delta )$ represents the measurement of the parameter modifications, which should be minimized for the attack implementation efficiency. 
In this paper, $\ell_0$ and $\ell_2$ norms are used as $D(\bm \delta )$ as follows,
\begin{equation}
D(\bm \delta ) = {\left\|\bm  \delta  \right\|_0} \quad  \text{or} \quad D(\bm \delta ) = {\left\|\bm  \delta  \right\|_2}.
\end{equation}
The ${\ell_0}$ norm of $\bm{\delta}$ measures the number of nonzero elements in $\bm{\delta}$ and therefore measures the number of modified parameters by the attack. Minimizing ${\ell_0}$ norm can make it easier to implement the attack in DNN systems, considering that the difficulty of parameter modifications in real systems relates to the number of modified parameters \cite{kim2014flipping}.
The ${\ell_2}$ norm of $\bm{\delta}$ denotes the standard Euclidean distance between the modified and original parameters, and therefore measures the magnitude of parameter modifications.  Minimizing ${\ell_2}$ norm can lead to minimal influence of the attack.

Minimizing the $\ell_0$ norm in the objective function is much harder than minimizing the $\ell_2$ norm, because the $\ell_0$ norm is non-differential. In this paper, the proposed ADMM framework enables both $\ell_0$ and $\ell_2$ norms in the objective function with only minor differences in the solution methods as specified in Sec. \ref{admm_solution}.

\subsection{Misclassification Requirements}

In \eqref{eq: original_problem}, $G(\bm \theta + \bm \delta, \mathcal X, \mathcal T, \mathcal L)$ denotes the misclassification requirements: 1) the first $S$ images ${\mathcal X_1} = \left\{ {{\bm x_i}|i = 1, \cdots ,S} \right\}$  should be classified as the target labels $\mathcal T$ instead of their correct labels, and 2) the classifications of the rest $R-S$ images ${\mathcal X_2} = \left\{ {\bm x_i|i = S+1, \cdots ,R} \right\}$ should remain unchanged as their correct labels.  

In the area of adversarial machine learning, the most effective objective function to specify that an input $\bm x$ should be labeled as $t$ is the following $g$ function \cite{carlini2017towards}:
\begin{equation}
g(\bm \theta + \bm \delta, t) = \max \left( {\mathop {\max }\limits_{j \ne {t}} (Z{{(\bm \theta  + \bm \delta ,{\bm x})}_j}) - Z{{\left( {\bm \theta  + \bm \delta ,{\bm x}} \right)}_{{t}}},0} \right)
\end{equation}
where $Z{{\left( {\bm \theta  + \bm \delta ,{\bm x}} \right)}_{{j}}}$ denotes the $j$-th element of the logits, i.e., the input to the softmax layer. 
The softmax layer is the last layer in the DNN model, which takes logits as input and generates the final probability distribution outputs.
The final outputs from the softmax layer are not utilized in the above $g$ function, because the final outputs are usually dominated by the most significant class in a well trained model and thus less effective during computation. 
The DNN chooses the label with the largest logit, that is, $j^*= \arg \mathop {\max }\limits_{j} \ Z{{(\bm \theta  + \bm \delta ,{\bm x})}_j}$. To enforce the input $\bm x$ is classified as label $t$, the logit of label $t$, $Z{{\left( {\bm \theta  + \bm \delta ,{\bm x}} \right)}_{{t}}}$, must be larger than all of the other logits, $\mathop {\max }\limits_{j \ne {t}} (Z{{(\bm \theta  + \bm \delta ,{\bm x})}_j})$.
Thus, $g(\bm \theta + \bm \delta, t)$ will achieve its minimal value if $\bm x$ is classified as label $t$.

From the above analysis, we propose the detailed form of $G$ as: 
\begin{equation}
G(\bm \theta + \bm \delta, \mathcal X, \mathcal T, \mathcal L)=  G_1(\bm \theta + \bm \delta, \mathcal X_1, \mathcal T) + G_2(\bm \theta + \bm \delta, \mathcal X_2, \mathcal L),
\end{equation}
where $G_1$ stands for the targeted misclassifications of ${\mathcal X_1}$ and $G_2$ denotes keeping classifications of ${\mathcal X_2}$ unchanged. $G_1$ and $G_2$ are:
\begin{align}
\small
& G_1(\bm \theta + \bm \delta, \mathcal X_1, \mathcal T)  
\nonumber \\
& = \sum\limits_{i = 1}^S {c_i} \cdot \max \left( {\mathop {\max }\limits_{j \ne t_i} (Z{{(\bm \theta  + \bm \delta ,{\bm x_i})}_j})  - Z{{( {\bm \theta  + \bm \delta ,{\bm x_i}} )}_{t_i}},0} \right),
\end{align}
\begin{align}
\small
&G_2(\bm \theta + \bm \delta, \mathcal X_2, \mathcal L) 
\nonumber \\ &=\sum\limits_{i = S + 1}^R {{c_i} \cdot \max \left( {\mathop {\max }\limits_{j \ne {l_i}} (Z{{(\bm \theta  + \bm \delta ,{\bm x_i})}_j}) - Z{{\left( {\bm \theta  + \bm \delta ,{\bm x_i}} \right)}_{{l_i}}},0} \right)} .
\end{align}
The $c_i$'s represent their relative importance to the measurement of modifications $D(\bm \delta)$.  $t_i$ represents the target label for the $i$-th image in the $S$ images. 
$G_1$ achieves its minimum value, when the labels of the first $S$ images are changed to their target labels $\mathcal T$. Similarly, $G_2$ obtains its minimum value when the classifications of the rest $R-S$ images are kept unchanged.


\section{General ADMM Solution Framework} \label{admm_solution}

We propose a solution framework based on ADMM to solve \eqref{eq: original_problem} for the fault sneaking attack. The framework is \emph{general} in that it can deal with both $\ell_0$ and $\ell_2$ norms as $D(\bm \delta)$.
ADMM was first introduced in the mid-1970s with roots in the 1950s
and becomes popular recently for large scale statistics and machine learning problems 
\cite{boyd2011distributed}. 
ADMM solves the problems in the form of a decomposition-alternating procedure, where the global problem is split into local subproblems first, and then the solutions to small local subproblems are coordinated to find a solution to the large global problem.
It has been proved in \cite{hong2017linear} that ADMM has at least the linear convergence rate, and it empirically converges in a few tens of iterations. 

\subsection{ADMM Reformulation}
As ADMM requires multiple variables for reducing the objective function in alternating directions, we introduce a new auxiliary variable $\bm z$ and \eqref{eq: original_problem} can now be reformulated as,
\begin{equation}
\begin{array}{l}
\mathop {\min }\limits_{\bm \delta ,\bm z} \quad D(\bm z) + G(\bm \theta  + \bm \delta ,\mathcal X, \mathcal T, \mathcal L),\\
{\rm{s}}{\rm{.t}}{\rm{.}}\quad \;\;\bm z=\bm \delta .
\end{array}
\end{equation}
The augmented Lagrangian function of the above problem is:
\begin{equation}
\small
{L_\rho }(\bm \delta ,\bm z,\bm u) = D(\bm z) + G(\bm \theta  + \bm \delta , \mathcal X, \mathcal T, \mathcal L) + {\bm u^T}(\bm z - \bm \delta) + \frac{\rho }{2}\left\| {\bm z - \bm \delta} \right\|_2^2.
\end{equation}
Applying the scaled form of ADMM by defining $\bm u=\rho \bm s$, we obtain
\begin{equation} \label{eq: Lagrangian}
\small
{L_\rho }(\bm \delta ,\bm z,\bm s) =D(\bm z) + G(\bm \theta  + \bm \delta ,\mathcal X, \mathcal T, \mathcal L) + \frac{\rho }{2}\left\| {\bm z - \bm \delta + \bm s} \right\|_2^2 - \frac{\rho }{2}\left\| \bm s \right\|_2^2.
\end{equation}

\subsection{ADMM Iterations}
ADMM optimizes problem \eqref{eq: Lagrangian} in iterations. Specifically, in the $k$-th iteration, the following steps are performed:
\begin{equation} \label{eq: z-problem}
{{\bm z}^{k + 1}} = \arg \mathop {\min }\limits_{\bm z} {L_\rho }({\bm \delta ^k},\bm z,{\bm s^k}),
\end{equation}
\begin{equation} \label{eq: d-problem}
{{\bm \delta} ^{k + 1}} = \arg \mathop {\min }\limits_{\bm \delta}  {L_\rho }(\bm \delta ,{\bm z^{k + 1}},{\bm s^k}),
\end{equation}
\begin{equation}  \label{eq: updates}
{\bm s^{k + 1}} = {\bm s^k} + {\bm z^{k + 1}} - {\bm \delta ^{k + 1}}.
\end{equation}
As demonstrated above, problem \eqref{eq: Lagrangian} is split into two subproblems, \eqref{eq: z-problem} and \eqref{eq: d-problem}  through ADMM. In \eqref{eq: z-problem}, the optimal solution ${\bm z}^{k + 1}$ is obtained by minimizing the  augmented Lagrangian function ${L_\rho }({\bm \delta ^k},\bm z,{\bm s^k})$ with fixed $\bm \delta ^k$ and $\bm s^k$. Similarly, \eqref{eq: d-problem} finds the optimal ${\bm \delta} ^{k + 1}$ to minimize $ {L_\rho }(\bm \delta ,{\bm z^{k + 1}},{\bm s^k})$ with fixed ${\bm z^{k + 1}}$ and $\bm s^k$. In \eqref{eq: updates}, we update $\bm s^{k + 1}$ with ${\bm z}^{k + 1}$ and ${\bm \delta} ^{k + 1}$. We can observe that ADMM updates the two arguments in an alternating fashion, where comes from the term \textit{alternating direction}. 

In the ADMM iterations, problems \eqref{eq: z-problem} and \eqref{eq: d-problem} are detailed as:
\begin{equation}\label{eq: z-problem-detailed}
\mathop {\min }\limits_{\bm z} \quad D(\bm z) + \frac{\rho }{2}\left\| { \bm z - \bm \delta^k  + \bm s^k} \right\|_2^2,
\end{equation}

\begin{equation}\label{eq: d-problem-detailed}
\mathop {\min }\limits_{\bm \delta}  \quad G(\bm \theta  + \bm \delta ,\mathcal X, \mathcal T, \mathcal L) + \frac{\rho }{2}\left\| {\bm z^{k+1} - \bm \delta  + \bm s^k} \right\|_2^2.
\end{equation}
The solutions to the two problems are specified 
as follows.

\subsection{z step}
In this step, we mainly solve \eqref{eq: z-problem-detailed}. The specific closed-form solution depends on the $D$ function ($\ell_0$ or $\ell_2$ norm).

\subsubsection{Solution for $\ell_0$ norm}
If the $D$ function takes the $\ell_0$ norm, \eqref{eq: z-problem-detailed} has the following form:
\begin{equation}
\mathop {\min }\limits_{\bm z} \quad {\left\| \bm z \right\|_0} + \frac{\rho }{2}\left\| {\bm z - \bm \delta^k  + \bm s^k} \right\|_2^2.
\end{equation}
The solution  can be obtained  elementwise \cite{parikh2014proximal} as
\begin{equation}
\bm z _i^{k+1} = \left\{ {\begin{array}{*{20}{c}}
{{{\left( {\bm \delta^k - \bm s^k} \right)}_i},}&{{\rm{if}}\;\left( {\bm \delta^k - \bm s^k} \right)_i^2 > \frac{2}{\rho }}\\
{0,}&{{\rm{otherwise}}}
\end{array}} \right..
\end{equation}

\subsubsection{Solution for $\ell_2$ norm}
If the $D$ function takes the $\ell_2$ norm, \eqref{eq: z-problem-detailed} has the following form:
\begin{equation}
\mathop {\min }\limits_{\bm z} \quad {\left\| \bm z \right\|_2} + \frac{\rho }{2}\left\| {\bm z - \bm \delta^k  + \bm s^k} \right\|_2^2 .
\end{equation}
 By `block soft thresholding' operator \cite{parikh2014proximal}, the solution is given by
\begin{equation}
{z^{k + 1}} = \left\{ {\begin{array}{*{20}{c}}
{\left( {1 - \frac{1}{{\rho {{\left\| {{\delta ^k} - {s^k}} \right\|}_2}}}} \right)\left( {{\delta ^k} - {s^k}} \right)}&{{\rm{if}}\;{{\left\| {{\delta ^k} - {s^k}} \right\|}_2} \ge \frac{1}{\rho }}\\
0&{{\rm{if}}\;{{\left\| {{\delta ^k} - {s^k}} \right\|}_2} < \frac{1}{\rho }}
\end{array}} \right..
\end{equation}

\subsection{$ \delta$ step}  \label{sec: d-step}
In this step, we mainly solve \eqref{eq: d-problem-detailed}. It can be rewritten as
\begin{equation}   \label{d-problem}
\mathop {\min }\limits_{\bm \delta}  \quad \sum\limits_{i = 1}^R {{g_i}(\bm \theta  +\bm \delta ,{\bm x_i})}  + \frac{\rho }{2}\left\| { \bm z^{k+1} - \bm \delta   +\bm  s^k} \right\|_2^2 ,
\end{equation}
where
\begin{equation}
{g_i}(\theta  + \delta ,{x_i}) = \left\{ \begin{array}{l}
{c_i}\cdot \max \left( {\mathop {\max }\limits_{j \ne {t_i}} \left( {Z{{\left( {\bm \theta  + \bm \delta ,{\bm x_i}} \right)}_j}} \right) - Z{{\left( {\bm \theta  + \bm \delta ,{\bm x_i}} \right)}_{{t_i}}},0} \right),\\
\quad \quad \quad \quad \quad \quad \quad \quad \quad \quad \quad \quad \quad \quad {\rm{if}}\;i \in \left[ {1,S} \right];\\
{c_i}\cdot \max \left( {\mathop {\max }\limits_{j \ne {l_i}} \left( {Z{{\left( {\bm \theta  + \bm \delta ,{\bm x_i}} \right)}_j}} \right) - Z{{\left( {\bm \theta  + \bm \delta ,{\bm x_i}} \right)}_{{l_i}}},0} \right),\\
\quad \quad \quad \quad \quad \quad \quad \quad \quad \quad \quad \quad \;\,{\rm{if}}\;i \in \left[ {S + 1,R} \right].
\end{array} \right.
\end{equation}
The $g_i$ function takes different forms according to the  $i$ value. If $i \in \left[ {1,S} \right]$, $g_i$ obtains its minimum value when the classification of $\bm x_i$ is changed to the target label $t_i$. If $i \in \left[ {S+1,R} \right]$, $g_i$ achieves its minimum when the classification is kept as the original label $l_i$.

Motivated by the linearized ADMM \cite[Sec.\,2.2]{gao2017first,liu2017linearized}, we replace the function $g_i$ with its first-order Taylor expansion plus a regularization term (known as Bregman divergence), 
$\nabla {g_i}(\bm \theta  + {\bm \delta ^k},{\bm x_i},l_i)(\bm \delta  - {\bm \delta ^k}) + \frac{1}{2}\left\| {\bm \delta  - {\bm \delta ^k}} \right\|_H^2$,
where $\bm H$ is a pre-defined positive definite matrix , and $ \| \bm x \|_{\bm H}^2 =  \bm x^T \bm H \bm x $. \eqref{eq: d-problem-detailed} can then be reformulated as:
\begin{align}
\mathop {\min }\limits_{\bm \delta}  \quad & \left( {\sum\limits_{i = 1}^R {\nabla {g_i}(\bm \theta  + {\bm \delta ^k},{\bm x_i})} } \right)(\bm \delta  - {\bm \delta ^k}) + \frac{R}{2}\left\| {\bm \delta  - {\bm \delta ^k}} \right\|_H^2  \nonumber \\ 
& + \frac{\rho }{2}\left\| {\bm z^{k+1} - \bm \delta  + \bm s^k} \right\|_2^2 .
\end{align}
Letting $\bm H = \alpha \bm I$, the solution can be obtained through
\begin{equation}
{\bm \delta ^{k+1}} = \frac{1}{{\alpha R + \rho }}\left( {\rho \left( {\bm z^{k+1} + \bm s^k} \right) + \alpha R{\bm \delta ^k} - \left( {\sum\limits_{i = 1}^R {\nabla {g_i}(\bm \theta  + {\bm \delta ^k},{\bm x_i})} } \right)} \right) .
\end{equation}

\section{Experimental Evaluations}
We demonstrate the experimental results of the proposed fault sneaking attack on two image classification datasets, MNIST \cite{Lecun1998gradient} and CIFAR-10 \cite{Krizhevsky2009learning}. 
We train two networks  for MNIST and CIFAR-10 datasets, respectively, sharing the same  network architecture with four convolutional layers, two max pooling layers, two fully connected layers and one softmax layer. 
They achieve 99.5\% accuracy on MNIST and 79.5\% accuracy on CIFAR-10, respectively, which are comparable to the state-of-the-arts. The experiments are conducted on machines with NVIDIA GTX 1080 TI GPUs.

\subsection{\textcolor{black}{Layer and Type of Parameters to Modify}}


\begin{table}
\vspace{-2mm}
\begin{center}
\caption{$\ell_0$ norm of DNN parameter modifications (i.e., the number of modified parameters) in different fully connected layers for MNIST.}
\label{tab: layers}
\scalebox{0.8}{
\begin{tabular}{l|c|c|c|c}
\toprule[1pt]
 & \multirow{2}{*}{Total Parameters} & \multicolumn{3}{c}{$\ell_0$ norm} \\ \cline{3-5}
 &  & S=1,R=1  & S=4,R=4  &  S=16,R=16 \\
\hline
The first FC layer & 205000 & 14016 &40649  & 120597 \\
The second FC layer & 40200 & 5390 &14086 & 34069 \\
The last FC layer & 2010 & 222 & 682 & 1755\\
\bottomrule[1pt]
\end{tabular}}
\end{center}
\end{table}

\begin{table}[tb]
\vspace{-2mm}
\begin{center}

\caption{$\ell_0$ norm and attack success rate when modifying different types of parameters in the last fully connected layer for MNIST.}

\label{tab: kernel_bias}
\scalebox{0.8}{
\begin{threeparttable}
\begin{tabular}{l|c|c|c|c}
\toprule[1pt]
  & S=1, R=1 & S=2, R=2  & S=4, R=4  &  S=8, R=8 \\
\hline
$\ell_0$ norm for weight params. & 236 & 458 &715 & 1644 \\
Success rate for weight params. & 100\% & 100\% &100\% & 100\% \\ \hline
$\ell_0$ norm for bias params.  & 2 & 4 & -\tnote{*} & -\tnote{*}\\
Success rate for bias params. & 100\% & 100\% & 0\% & 0\%\\
\bottomrule[1pt]
\end{tabular}
\begin{tablenotes}
    \small
    \item[*] There is no need to show the $\ell_0$ norm if it can not succeed.
\end{tablenotes}
\end{threeparttable}}
\end{center}
\end{table}

The DNN model used has three fully connected (FC) layers. We modify the parameters in different FC layers. We show the $\ell_0$ norm (i.e., the number of parameter modifications) achieved by the fault sneaking attack when we modify each FC layer in Table \ref{tab: layers}.
We observe that more parameters are needed to be modified with increasing $S$ and $R$. Besides, changing the last FC layer requires fewer parameter modifications compared with the first or second FC layer. 
The reason is that the last FC layer has more direct influence on the output, leading to smaller number of modifications by the fault sneaking attack. 
Therefore, in the following experiments, we focus on modifying only the last FC layer parameters.


Next we determine the type of parameters to modify that is more effective to implement the fault sneaking attack.
In the FC layer, the output depends on 
the weights $\bm W$ and the biases $\bm b$, that is, 
$FC(\bm x^\prime) = \bm W \bm x^\prime + \bm b,$
where $\bm x^\prime$ is the input of the layer. As we can see, the bias parameters are more directly related to the output than the weight parameters.
We show the $\ell_0$ norm and the attack success rate if we only modify the weight parameters or the bias parameters in the last FC layer in Table \ref{tab: kernel_bias}. 
As the bias parameters are more directly related to the output, it usually needs to change fewer bias parameters to achieve the same attack objective. However, only changing bias parameters has very limited capability which can only lead to the misclassification of 1 or 2 images.  As observed from Table \ref{tab: kernel_bias}, changing the classification of 4 or more images would be beyond the capability of modifying bias parameters only.
\textcolor{black}{This demonstrates the limitation of the single bias attack (SBA) scheme in \cite{liu2017fault}, which only modifies the bias to misclassify only one image. Also we find that SBA can not be extended to solve the case of multiple images with multiple target labels. }
Considering the limitation of only modifying bias parameters, we choose to perturb both the weight and bias parameters in the following experiments. 


\subsection{$\ell_0$ Norm of Parameter Modifications}

\begin{figure} 
     \vspace{-2mm}
    \centering
    \includegraphics[width=0.32 \textwidth]{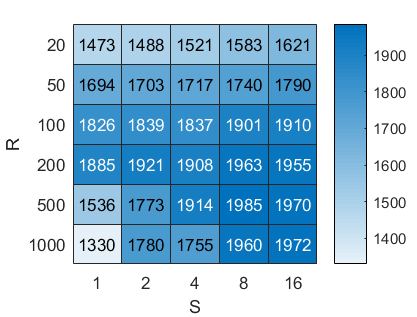}
    \vspace{-3mm}
    \caption{$\ell_0$  norm of DNN parameter modifications in the last fully connected layer for MNIST.}
    \label{fig: l0mnist}
     \vspace{-4mm}
\end{figure}

\begin{figure}
    \centering
    \includegraphics[width=0.32 \textwidth]{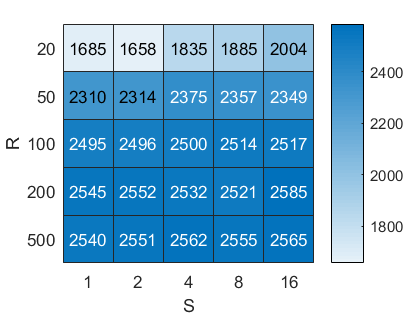}
    \vspace{-3mm}
    \caption{$\ell_0$ norm of DNN parameter modifications in the last fully connected layer for CIFAR-10.}
    \label{fig: l0cifar}
    \vspace{-4mm}
\end{figure}

We demonstrate the number of parameter modifications, i.e., the $\ell_0$ norm, by the fault sneaking attack in this section. As observed from Fig. \ref{fig: l0mnist} and \ref{fig: l0cifar}, for the same $R$, the $\ell_0$ norm of parameter modifications keeps increasing as $S$ increases since more parameters need to be modified to change the classifications of more images into their target labels. 
We have an interesting finding  that when $S$ is in the range of $\{1,2,4\}$, the $\ell_0$ norm tends to be smaller as $R$ increases from 200 to 1000 for MNIST. 
The reason is that larger $R$ means   the labels of more images ($R-S$) need to be kept unchanged, then the modified model should be more similar to the original model and therefore fewer modifications are required. 

We also notice that this phenomenon disappears when $S$ is larger than 8 for MNIST or for CIFAR-10. Considering the  99.5\% and 79.5\%  accuracy on MNIST and CIFAR-10, we believe the disappearance is related to the DNN model capability. When $S$ is small on MNIST, the DNN model is able to hide a small number  of misclassifications by modifying only a few parameters of the last FC layer. However, when $S$ is relatively large, it is not that easy to hide so many misclassifications and the fault sneaking attack has to perturb almost all parameters in the last FC layer without extra ability to spare. The reason for CIFAR-10 is similar since the capability of the model for CIFAR-10 is limited, with only 79.5\% accuracy.

\subsection{Comparison of $\ell_0$ and $\ell_2$ based Attacks}
In problem \eqref{eq: z-problem}, the $\ell_0$ or $\ell_2$ norm can be minimized, leading to the corresponding  $\ell_0$ or $\ell_2$ based fault sneaking attacks. 
Table \ref{tab: l0_l2} compares the $\ell_0$ and $\ell_2$ norms of the $\ell_0$ and $\ell_2$ based attacks for various configurations. 
\begin{table}[tb]
\vspace{-3mm}
\begin{center}
\caption{$\ell_0$ and $\ell_2$ norms of DNN parameter modifications in the last fully connected layer for the $\ell_0$ and $\ell_2$ based attacks for MNIST.}
\label{tab: l0_l2}
\scalebox{0.85}{
\begin{threeparttable}
\begin{tabular}{l|c|c|c|c|c|c}
\toprule[1pt]
& \multicolumn{2}{c|}{S=1, R=10} & \multicolumn{2}{c|}{S=5, R=10}  & \multicolumn{2}{c}{S=5, R=20}  \\
\cline{2-7}
 & $\ell_0$ norm & $\ell_2$ norm & $\ell_0$ norm &$\ell_2$ norm & $\ell_0$ norm  & $\ell_2$ norm\\
\hline
$\ell_0$ attack & 1026 & 863 & 1208 & 804 & 1606 & 498\\
$\ell_2$ attack & 1431 & 393 & 1432 & 344  & 1964 & 226 \\
\bottomrule[1pt]
\end{tabular}
\end{threeparttable}}
\end{center}
\vspace{-3mm}
\end{table}
As seen from Table \ref{tab: l0_l2}, the $\ell_0$ based attack achieves smaller $\ell_0$ norm than the $\ell_2$ based attack with larger $\ell_2$ norm, due to the reason that the $\ell_2$ based attack tries to minimize the Euclidean distance between the perturbed and original model without considering the number of parameter modifications.
 
\subsection{Test Accuracy after Parameter Modification}

As the fault sneaking attack perturbs the DNN parameters to satisfy specific attack requirements, it is important to measure the influence of the attack beyond the required objective. In the problem formulation, we try to reduce the influence of fault sneaking attack by enforcing the rest $R-S$ images to have unchanged classifications. In Table \ref{tab: accuracy_afterwards}, we show the test accuracy on the whole testing datasets for MNIST and CIFAR-10 after perturbing the model.  

\begin{table}[tb]
\begin{center}
\caption{Test accuracy after DNN parameter modifications for MNIST and CIFAR.}
\label{tab: accuracy_afterwards}
\scalebox{1.0}{
\begin{tabular}{l|l|c|c|c|c|c}
\toprule[1pt]
 Dataset & Test Acc. & S=1 & S=2 & S=4  & S=8 & S=16 \\
\hline
 \multirow{5}{*}{MNIST} & R=50 & 85.2\% & 73.1\% & 64.7\% & 37.4\% & 29.7\% \\
  & R=100 & 96.9\% & 86.6\% & 81.3\% & 76.1\% & 65.2\% \\
  & R=200  & 96.7\% & 96.1\% & 95.4\% & 93.2\% & 92.6\%\\
  & R=500 & 98.6\% & 98.5\% & 97.8\% & 96.9\% & 95.9\%\\
  & R=1000 & 98.7\% & 97.9\% & 98.1\% & 96.8\% & 96.9\%\\
  \hline
\multirow{5}{*}{CIFAR} & R=50 & 57.7\% & 52.9\% & 44.9\% & 26.2\% & 18.3\% \\
  & R=100 & 67.5\% & 68.7\% & 55.8\%& 42.5\% & 31.5\% \\
  & R=200  & 72.3\% & 67.6\% & 69.6\% & 57.2\% & 35.4\%\\
  & R=500 & 78.5\% & 77.4\% & 76.2\% & 74.5\% & 73.2\%\\
  & R=1000 & 78.5\% &	78.2\% &	77.5\% &	77.9\%	& 76.4\%\\
\bottomrule[1pt]
\end{tabular}}
\end{center}
\end{table}


The test accuracy of the original model is 99.5\% for MNIST and 79.5\% for CIFAR. As observed from Table \ref{tab: accuracy_afterwards}, with fixed $R$,
the test accuracy on the modified model decreases as $S$ increases.
This demonstrates that as a nature outcome, changing parameters to misclassify certain  images may downgrade the overall accuracy performance of the model. In the case of $S=16$ and $R=50$, the test accuracy drops from 99.5\% to 29.7\% for MNIST and from 79.5\% to 18.3\% for CIFAR. 
However, we observe that as $R$ increases, the test accuracy keeps increasing for fixed $S$. It demonstrates that keeping the labels of the $R-S$ images unchanged helps to stabilize the model and reduce the influence of changing the labels of the $S$ images. In the case of $S=16$, if $R$ is increased from 50 to 1000, the test accuracy on the 10,000 test images increases from 29.7\% to 96.9\% for MNIST and from 18.3\% to 76.4\% for CIFAR. 
The fault sneaking attack can achieve classification accuracy as high as 98.7\% and 78.5\% in the case of $S=1$ and $R=1000$ for MNIST and CIFAR, which only degrades the accuracy by 0.8 percent and 1.0 percent respectively, from the original models. 
\textcolor{black}{Note that under the same assumption of misclassifying only one image, \cite{liu2017fault} degrades the accuracy by 3.86 percent and 2.35 percent, respectively, for MNIST and CIFAR in the best case.
Compared with \cite{liu2017fault}, the proposed attack achieves a great improvement to reduce the influence of model perturbation.}



\begin{figure}
\vspace{-4mm}
\begin{minipage}[t]{0.24\textwidth}
\centering
\includegraphics[width=1.05\textwidth]{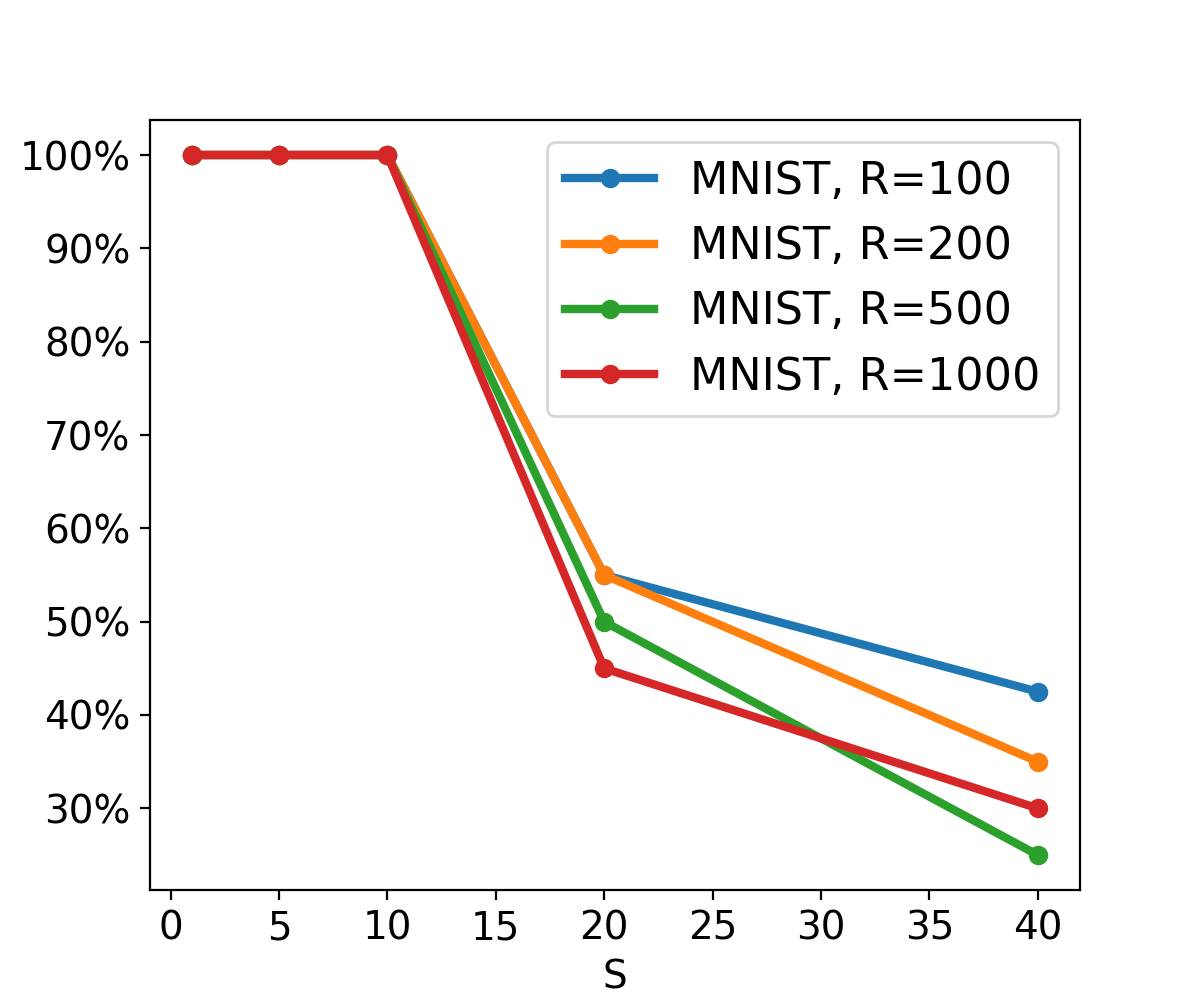}
\end{minipage}%
\begin{minipage}[t]{0.24\textwidth}
\centering
\includegraphics[width=1.05\textwidth]{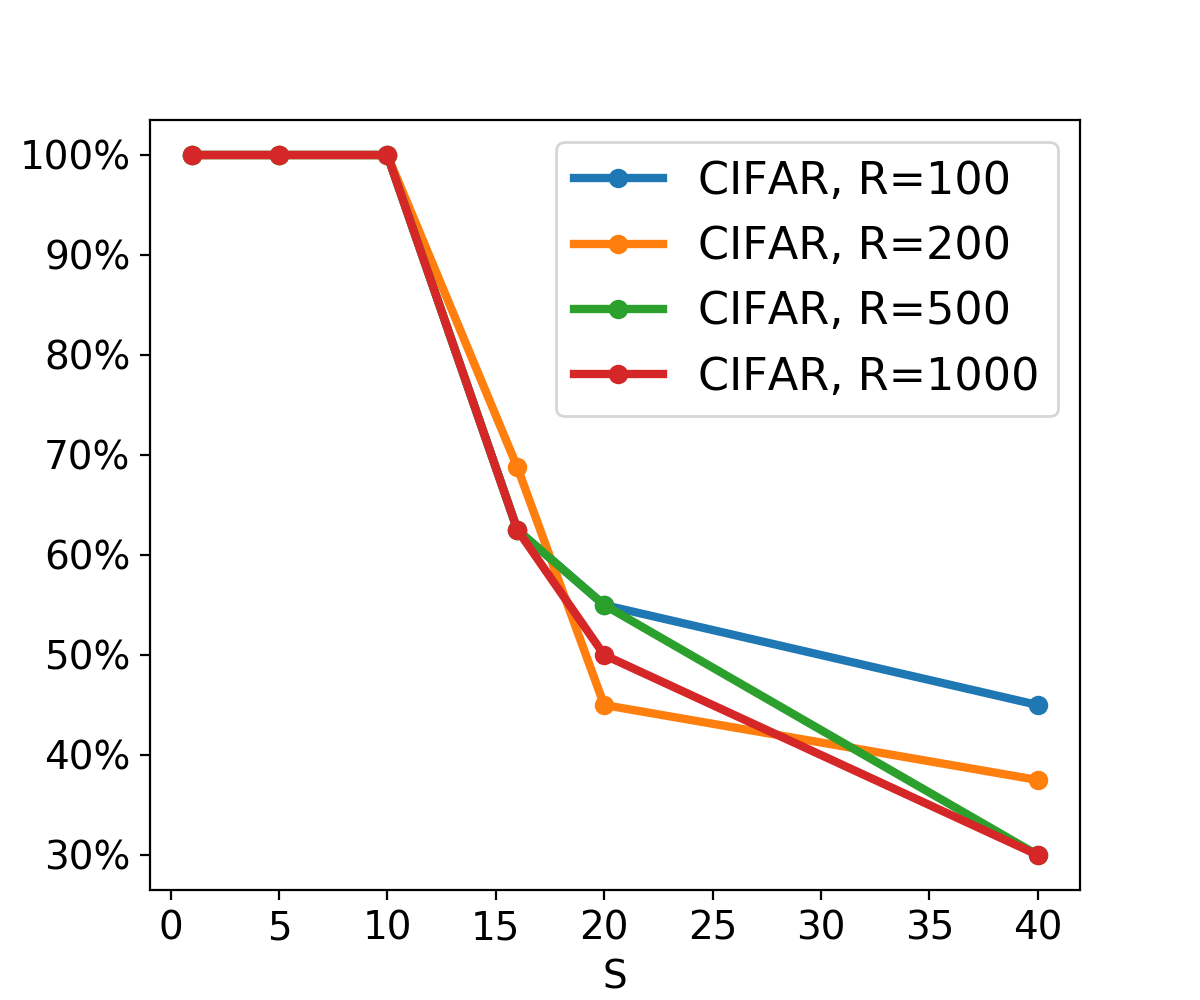}
\end{minipage}
\caption{Fault sneaking attack success rate of the $S$ images  after DNN parameter modifications for MNIST and CIFAR.}
\label{fig: MNIST_CIFAR_SR}
\vspace{-4mm}
\end{figure}

\subsection{Tolerance for Sneaking Faults}
One objective of fault sneaking attack is to hide faults by perturbing the DNN parameters.
In the experiments, we found that in the case of large $S$, not all of the $S$ images are changed to their target labels successfully. We define the success rate of the $S$ images as the percentage of images successfully changed their labels to the target labels within the $S$ images. We show the success rate of the $S$ images with various $S$ and $R$ configurations  in Fig. \ref{fig: MNIST_CIFAR_SR}. We observe that the success rate  keeps almost 100\% if  $S$ is smaller than 10. When $S$ is larger than 10, the success rate would drop as $S$ increases. 
Besides, the  number of successful injected faults in $S$ 
is usually around 10 for different configuration of $S$. This demonstrates a limitation of changing the classifications of certain images by modifying DNN parameters. The DNN model has a tolerance for the sneaking faults - 10 successful misclassifications by modifying the last FC layer.

\section{Conclusion}
In this paper, we propose fault sneaking attack to mislead the DNN by modifying model parameters.  The $\ell_0$  and $\ell_2$ norms  are minimized by the general framework with constraints to keep the classification of other images unchanged.
The experimental evaluations demonstrate that the ADMM based framework can implement the attacks  stealthily and efficiently with negligible test accuracy loss. 

\section{Acknowledgement}
This work is supported by Air Force Research Laboratory FA8750-18-2-0058, and U.S. Office of Naval Research.

\bibliographystyle{ieeetr}
\bibliography{reference}

\end{document}